\def\year{2022}\relax
\documentclass[letterpaper]{article} 
\usepackage{aaai22}  
\usepackage{times}  
\usepackage{helvet}  
\usepackage{courier}  
\usepackage[hyphens]{url}  
\usepackage{graphicx} 
\usepackage{natbib}  
\usepackage{caption} 
\DeclareCaptionStyle{ruled}{labelfont=normalfont,labelsep=colon,strut=off} 
\usepackage{algorithm}
\usepackage{algorithmic}
\usepackage{multirow}
\usepackage{adjustbox}
\usepackage{amssymb,amsmath,amsthm}
\newtheorem{theorem}{Theorem}
\newtheorem{lemma}[theorem]{Lemma}
\usepackage{newfloat}
\usepackage{listings}
\floatstyle{ruled}
\newfloat{listing}{tb}{lst}{}
\floatname{listing}{Listing}
\title{Disentangled Spatiotemporal Graph Generative Models}
\author{
    Yuanqi Du\textsuperscript{\rm 1}, Xiaojie Guo\textsuperscript{\rm 2 \rm*}, Hengning Cao\textsuperscript{\rm 1}, Yanfang Ye\textsuperscript{\rm 3}, Liang Zhao\textsuperscript{\rm 4 \rm \dag}\\
}
\affiliations{
    \textsuperscript{\rm 1}George Mason University, Fairfax, US\\ 
    \textsuperscript{\rm 2}JD.COM Silicon Valley Research Center, Mountain View, CA, US\\\textsuperscript{\rm 3}University of Notre Dame, Notre Dame, US\\ \textsuperscript{\rm 4}Emory University, Atlanta, US\\
    \textsuperscript{\rm *}Equally contributing as first author\\
    \textsuperscript{\rm \dag}The corresponding author\\
}
\usepackage{bibentry}

\begin{document}

\maketitle

\begin{abstract}
Spatiotemporal graph represents a crucial data structure where the nodes and edges are embedded in a geometric space and can evolve dynamically over time. Nowadays, spatiotemporal graph data is becoming increasingly popular and important, ranging from microscale (e.g. protein folding), to middle-scale (e.g. dynamic functional connectivity), to macro-scale (e.g. human mobility network). Although disentangling and understanding the correlations among spatial, temporal, and graph aspects have been a long-standing key topic in network science, they typically rely on network processing hypothesized by human knowledge. This usually fit well towards the graph properties which can be predefined, but cannot do well for the most cases, especially for many key domains where the human has yet very limited knowledge such as protein folding and biological neuronal networks. In this paper, we aim at pushing forward the modeling and understanding of spatiotemporal graphs via new disentangled deep generative models. Specifically, a new Bayesian model is proposed that factorizes spatiotemporal graphs into spatial, temporal, and graph factors as well as the factors that explain the interplay among them. A variational objective function and new mutual information thresholding algorithms driven by information bottleneck theory have been proposed to maximize the disentanglement among the factors with theoretical guarantees. Qualitative and quantitative experiments on both synthetic and real-world datasets demonstrate the superiority of the proposed model over the state-of-the-arts by up to 69.2\% for graph generation and 41.5\% for interpretability.
\end{abstract}

\section{Introduction}
\label{sec:intro}

\begin{figure}[htbp]
\centering
\includegraphics[width=0.45\textwidth]{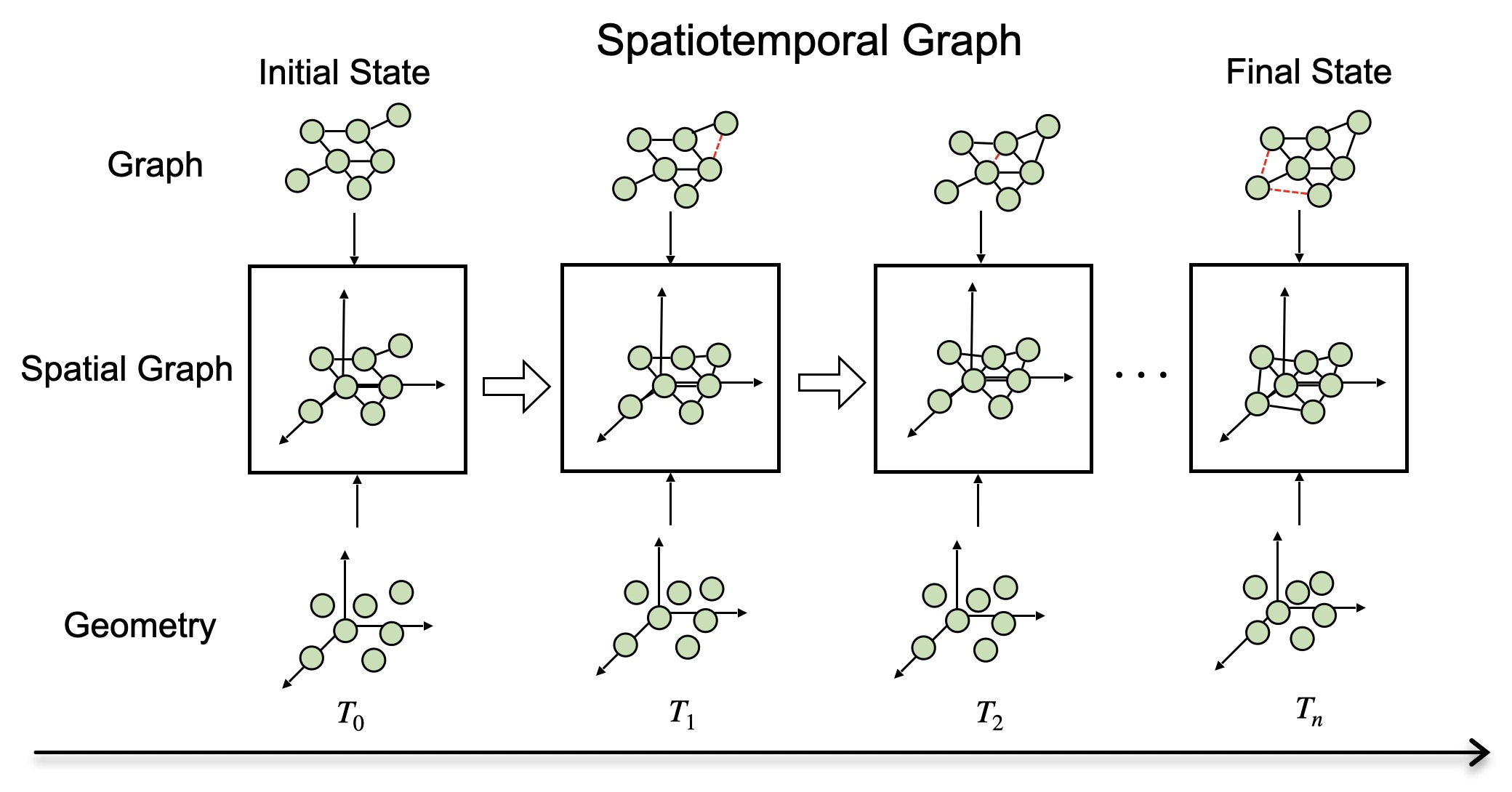}
\caption{Spatiotemporal graphs represent crucial data structures where the nodes and edges are embedded in a geometric space and their attribute values can evolve dynamically over time. Each column represents the formulation of a snapshot of a spatiotemporal graph.}
\label{fig:mot}
\end{figure}

There are two major directions on graph learning research in machine learning: 1) graph representation learning~\cite{kipf2016semi,velivckovic2017graph,hamilton2017inductive}, which aims at encoding graph
structural information into (low-dimensional) vector space, and 2) graph generation~\cite{you2018graphrnn,simonovsky2018graphvae}, which reversely aims at constructing a graph-structured data from low-dimensional space containing the graph generation rules or distribution. In this paper, we focus on the second direction, specifically for spatiotemporal graphs. Spatiotemporal graph represents a vital data structure where the nodes and edges are embedded and evolve in a geometric space, as shown in Fig. \ref{fig:mot}. Nowadays, spatiotemporal graph data is becoming increasingly popular and important, ranging from epidemic, transportation to biological network modeling \cite{dye2003modeling,ingraham2021generative,stopher1975urban,teng1985comparison,guo2020generating,fu2020core,du2020interpretable,xu2021opencda,xu2021opv2v,rahman2021generative,fu2021mimosa,zhao2021event,fu2021differentiable}. For example, the epidemic spreading network and the protein folding process can both be represented as spatiotemporal graphs, respectively.
Spatiotemporal graphs cannot be modeled using either the spatial, graph, or temporal information individually, but require the simultaneous characterization
of both the data and their interactions, which results in various patterns \cite{barthelemy2011spatial}. Spatial and graph aspects of information are usually correlated. For example, geographically nearby people tend to befriend in a social network. A pair of atoms that are very close in space tend to have a bond. Moreover, the above interplay between spatial and graph aspects is a dynamic process, thus, the consideration in time aspect is inevitable for a comprehensive modeling. Recently, although spatiotemporal graph deep learning has stimulated a surge of research for graph representation learning \cite{cui2019traffic,wu2019graph,yu2017spatio,khodayar2019convolutional,roy2021unified,wu2019graph,yu2017spatio,fu2021probabilistic}, however, deep generative models for spatiotemporal graph have not been well explored.

Modeling and understanding the generative process of spatiotemporal graphs are a long-lasting research topic in domains such as graph theory and network science.
Traditional methods usually extend and integrate network models in spatial networks (e.g., protein and molecule structures) and temporal graphs (e.g., traffic networks and epidemic spreading networks) into spatiotemporal graphs which captures some predefined properties of a graph, e.g., degree distribution, structure of community, clustering patterns. 
However, these models heavily rely on the predefined network process and rich knowledge of the graph properties, while the network
properties and generation principles always remain unknown in the real-world applications, such as models that explain the mechanisms of mental diseases in brain networks during an activity of human beings and protein structure folding. Another line of research works is computational simulation models of spatiotemporal graphs customized for specific applications such as epidemics, brain simulator, and transportation simulation \cite{dye2003modeling,stopher1975urban,teng1985comparison}. However, they are domain-specific with enormously detailed prior knowledge involved. This motivates us to propose the spatiotemporal graph models which can automatically learn the underlying spatial, temporal, and graph processes as well as their interplay.


Recent advanced deep generative models, such as variational auto-encoders (GraphVAE) \cite{kipf2016variational}, have made important progress towards
modeling and understanding each of the three components in spatiotemporal graphs, including graph data (e.g., citation networks) \cite{kipf2016semi}, spatial networks (e.g., proteins) \cite{ingraham2021generative} and temporal networks (e.g., traffic networks) \cite{zhang2020tg,zhou2020data}, respectively. However, none of the work is designed for spatiotemporal graphs which cannot be effectively modeled by simply integrating the existing techniques, due to several significant challenges: 
(1) \textbf{It is difficult to build a generic spatiotemporal graph generative framework to model spatial, graph, and temporal jointly}. This framework needs to not only capture the intrinsic generic properties shared across all types of spatiotemporal graphs, but also can automatically learn network models that tailor well for specific applications. 
(2) \textbf{It is difficult to effectively distinguish and capture the correlation among the spatial, temporal, and graph information}. As shown in Figure~\ref{fig:mot}, in spatiotemporal graphs, some network information are entangled, such as the only spatial-related information (Row 2) and only graph-related information (Row 3), while some spatial and graph properties are correlated (Row 1). (3) \textbf{It is difficult to theoretically ensure the disentanglement among the independent patterns}. In light of model interpretability, disentangling the independent patterns can help better control the properties of the generated spatialtemporal network. Current methods are limited in handling the disentanglement of spatial, temporal, and graph independent properties or factors.

To solve the aforementioned challenges, we propose, to the best of our knowledge, the first generic deep generative model framework that models and disentangles spatiotemporal graph data. Specifically, we first propose a novel deep Bayesian network that factorizes spatiotemporal graphs into the time-variant, time-invariant, spatial-graph joint, and independent factors based on inductive bias~\cite{liu2021physics}. A new objective driven by information-bottleneck theory has been proposed that can maximize the disentanglement of different factors as well as latent variables inside each factor, with theoretical guarantees. To optimize this objective function, a novel information-iterative-thresholding algorithm has been proposed to jointly optimize the objective and optimize its hyperparameters on information bottlenecks with theoretical analysis on optimal conditions. Extensive quantitative and qualitative experiments on two synthetic and two real-world datasets show the superiority of our proposed model over the state-of-the-art graph generative models by up to $69.2\%$ for spatiotemporal graph generation and $41.5\%$ for interpretability.

\begin{figure*}[htb]
    \centering
    \includegraphics[width=\textwidth]{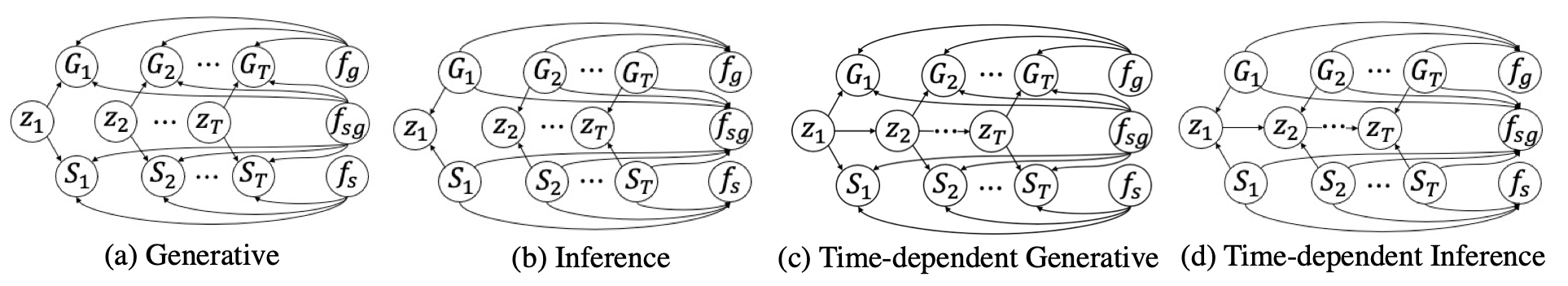}
    \caption{Graphical illustration of the proposed models. (a) The Bayesian network of the proposed probabilistic distribution of spatiotemporal graphs. (b) The approximate inference process of the posterior of latent variables, with conditional independence assumption across time snapshots. (c) The Bayesian network of the alternative probabilistic distribution of spatiotemporal graphs, with dependence assumption across time snapshots. (d) The alternative approximate inference model of the posterior of the proposed model, with dependence assumption across time snapshots.}
    \label{fig:pgm}
\end{figure*}


\section{Related Work}
\label{sec:relatedwork}

\textbf{Spatiotemporal Graph Deep Representation Learning}. This domain benefits a lot from the deep representation learning techniques for images, sequence, and network data, such as convLSTM. Currently, the most well-developed field is spatiotemporal forecasting in traffic networks \cite{cui2019traffic,wu2019graph,yu2017spatio,khodayar2019convolutional,roy2021unified,wu2019graph,yu2017spatio}. Other works \cite{wang2019molecule,zhang2020tg} either study spatial network data or temporal graph data. For example, \cite{wang2019molecule} considers the spatial locations of nodes, but not capable of generating temporal features. \cite{zhang2020tg} models the temporal features explicitly but do not include spatial features. 

\textbf{Spatiotemporal Graph Generation} is to generate diverse spatiotemporal graphs, motivated by network design and interpretation \cite{barthelemy2011spatial}. Traditional spatiotemporal network generation largely relies on human-defined heuristics/prior knowledge about the network being modeled, such as epidemic modeling, transportation modeling, protein modeling, etc. \cite{stopher1975urban,dye2003modeling,teng1985comparison}, which however almost pose no generalizability to other domain~\cite{dahiyat1997novo}. Another line of models are based on prescribed structural assumptions, such as probabilistic models \cite{barabasi1999mean}, configuration models \cite{bender1978asymptotic}, and stochastic block models \cite{xu2014dynamic}. These prescribed models aim to model some predefined graph properties, e.g. community structures, clustering patterns, etc, which, however, are not sufficient for real-world graph datasets where the prescribed rules are unkown~\cite{holme2015modern,rozenshtein2019mining}. Deep generative models have rarely been used to tackle the spatiotemporal graph generation problem \cite{zhang2019stggan}.

\textbf{Deep Generative Models on Graphs}.
Deep generative models achieve great success in computer vision, natural language processing, etc. Recently, increasing attention has been attracted to model graph-structured data by deep generative model. Work in \cite{simonovsky2018graphvae,you2018graphrnn,wang2018graphgan,shi2020graphaf,guo2020property,du2021equivariant} transfers popular deep generative models, such as GANs, VAEs, RNNs, Flow-based models, etc., into graph-structured data to model proteins, molecules, etc. and show promising results. For example, graphRNN utilizes an auto-regressive generative model to generate a sequence of nodes and edges by an LSTM model \cite{you2018graphrnn}. Graphite \cite{grover2019graphite} and VGAE \cite{kipf2016variational} focus on the node-level embedding and form edges between each pair of nodes to generate a graph. GraphVAE \cite{kipf2016variational} represents each graph by its edge feature (i.e. adjacency matrix) and node features, and utilizes an VAE model to learn the distribution of the graphs conditioned on the latent representation at the graph-level.



\textbf{Disentangled Representation Learning} 
Disentanglement learning aims to learn a disentangled representation that keeps latent variables separate and interpretable for the variations in the data. It has been shown that such disentangled representation improves the robustness against the adversarial attack and increases the generalizability of the model \cite{alemi2016deep}. This inspires much work in VAE to study how to disentangle the latent representations which expose real-world semantic factors and the solutions include adding, removing, or altering the weight of the objective term in the generative tasks \cite{chen2018isolating,kim2018disentangling,kumar2017variational,lopez2018information,zhao2017infovae}. Recently, the idea to learn a disentangled representation has also been applied to graph generative models. \cite{guo2020interpretable,ma2019disentangled} disentangles the latent variables that expose semantic factors of the nodes, edges, and the graphs. However, learning a disentangled representation for spatiotemporal graph modeling remains largely unexplored. 

\section{Methodology}
\label{sec:obj}

\subsection{Problem Formulation}
A spatiotemporal graph is defined as $(S_{1:T}, G_{1:T})$, where $T$ represents number of time frames of the spatiotemporal graphs, and $S_{1:T}=\{S_1,S_2,...S_T\}, G_{1:T} = \{G_1,...G_T\}$. $S_{t}=(\mathcal{V}_t, C_t)$ represents the geometric information of $t$-th snapshot of a spatiotemporal graph, where $\mathcal{V}_t $ denotes a set of $N$ nodes and $C_t \in\mathbb{R}^{N\times 3}$ denotes 3D geometric information. $G_t=(\mathcal{V}_t, \mathcal{E}_t,X_t, E_t)$ represents the graph information of $t$-th snapshot, where $\mathcal{E}_t \subseteq \mathcal{V}_t \times \mathcal{V}_t$ is the set of edges. $E_t\in\mathbb{R}^{N\times N \times K}$ refers to the edge weights or adjacent matrix, and $K$ refers to the edge feature dimension. $X_t\in\mathbb{R}^{N\times M}$ denotes the node feature and $M$ is the length of the node feature vector. It is worth noting that the geometric information $S_t$ can not be simply treated as part of node features since this type of representation cannot capture and maintain some properties, such as translation- and rotation-invariances~\cite{fuchs2020se}.

This paper aims at proposing a generic data-driven framework for modeling spatiotemporal graphs, under fundamental, necessary factors. First, for any spatiotemporal graphs, there could be patterns that are time-variant and time-invariant. While time-invariant, spatial and graph information could either be correlated or independent, hence it is important to distinguish and capture these different semantic factors via different latent variables. More concretely, the goal is to learn a posterior $p(S_{1:T}, G_{1:T}|Z,F)$ of the spatiotemporal graphs given four groups of generative latent variables $Z=z_{1:T}\in \mathbb{R}^{L_1}$ for time-variant features and $F=(f_s\in \mathbb{R}^{L_2},f_g\in \mathbb{R}^{L_3},f_{sg}\in \mathbb{R}^{L_4})$ for time-invariant features, where we need to capture and disentangle time-variant factors $z_{1:T}$, time-invariant geometric factors $f_{s}$, graph factors $f_{g}$ and spatial-graph joint factors $f_{sg}$. $L_1$, $L_2$, $L_3$, and $L_4$ are the number of variables in each group of factors, respectively. 
The encoding and generative process of our proposed \textbf{S}patio\textbf{T}emporal \textbf{G}raph \textbf{D}isentangled \textbf{V}ariational \textbf{A}uto-\textbf{E}ncoder (STGD-VAE) model is illustrated in Fig. \ref{fig:pgm}(a) and Fig. \ref{fig:pgm}(b). Another implementation of the proposed model following the common time-dependency, namely, STGD-VAE-Dep is illustrated in Fig. \ref{fig:pgm}(c) and Fig. \ref{fig:pgm}(d), and detailed in Supplementary Material.




\subsection{The Objective on Spatiotemporal Graph Generative Modeling}
To learn the conditional probability $p(S_{1:T},G_{1:T}|z_{1:T},F)$, it is equal to maximizing the marginal likelihood of the observed spatiotemporal graph sequence $(S_{1:T},G_{1:T})$ in expectation over the distribution of the latent representation as
     $ \mathbb{E}_{p_\theta(z_{1:T},F)}p_\theta(S_{1:T},G_{1:T}|z_{1:T}, F)$.
The prior distribution of the latent spaces is described as $p(z_{1:T}, F)$ with the observation of a spatiotemporal graph sequence $(S_{1:T}, G_{1:T})$, which, however, is intractable. Therefore, a variational objective is proposed to tackle this problem, where the posterior distribution is approximated by another distribution $q_{\phi}(z_{1:T},F|S_{1:T}, G_{1:T})$.
The objective can be written as minimizing the Kullback-Leibler Divergence (KLD) between the true prior distribution and the approximate posterior distribution. In order to encourage this disentanglement property of $q_{\phi}(z_{1:T},F|S_{1:T}, G_{1:T})$, we introduce a constraint by trying to match the inferred posterior configurations of the latent factors to the prior $p(z_{1:T},f_{s}, f_{g}, f_{sg})$. This can be achieved if we set each prior to be an isotropic unit Gaussian, i.e., $\mathcal{N}(\textbf{0},\textbf{1})$, leading to a constrained optimization problem as:
\begin{align}
    &\max_{\theta,\phi} \quad \mathbb{E}_{S_{1:T}, G_{1:T}\sim \mathcal{D}}[\mathbb{E}_{q_{\phi}(z_{1:T}, F|S_{1:T}, G_{1:T})} \nonumber\\
    &\qquad\qquad\qquad\qquad\qquad\qquad\log p_{\theta}(S_{1:T}, G_{1:T}|z_{1:T},F)]   \nonumber\\
    &\textrm{s.t.}D_{KL}(q_{\phi}(z_{1:T},F|S_{1:T}, G_{1:T})||p(z_{1:T},F))<I
    \label{infer1}
\end{align}
where $\mathcal{D}$ refers to the observed dataset of the spatiotemporal graphs and $I$ specifies the information that flows via the latent representation.

The above objective can be further decomposed for simple estimation and implementation of each component based on different pre-defined dependence and independence assumptions in the problem formulation, as stated in ~Lemma~\ref{lemma1}.

\begin{lemma}
\label{lemma1}
Given the assumption that: (1) $S_{1:T}\perp G_{1:T}|(z_{1:t}, F)$; (2) $S_{1:T}\perp f_g$ and $G_{1:T}\perp f_s$; (3) $G_i \perp G_j|(z_i,z_j,f_g,f_{sg})$ and $S_i \perp S_j|(z_t,z_k,f_s,f_{sg})$; (4) $z_{1:T} \perp (f_s, f_g, f_{sg})$, and $z_1 \perp z_2 \cdots \perp z_{T}$,  the objective of spatiotemoral graph generation can be expressed as
\begin{align}\nonumber
    \max_{\theta,\phi} \quad&\mathbb{E}_{S_{1:T}, G_{1:T}\sim \mathcal{D}}\mathbb{E}_{q_{\phi}(z_{1:t},F|G_{1:T},S_{1:T})}\\\nonumber
    &\sum\nolimits_{t=1}^{T}[\log p_{\theta} (G_t|z_t,f_{g},f_{sg})+ \log p_{\theta}(S_t|z_t,f_{s},f_{sg})]  \\\nonumber
    \textrm{s.t.} &\quad \sum\nolimits_{t=1}^{T} D_{KL}( q_{\phi}(z_t|G_t,S_t)||p(z_t))<I_{t}\\\nonumber
    & \quad D_{KL}(q_{\phi}(f_{g}|G_{1:T})||p(f_{g}))<I_{g}\\\nonumber
    & \quad D_{KL}(q_{\phi}(f_{s}|S_{1:T})||p(f_{s}))<I_{s} \\
    & \quad D_{KL}(q_{\phi}(f_{sg}|S_{1:T}, G_{1:T})||p(f_{sg}))<I_{sg}
    \label{infer5}
\end{align}
\end{lemma}
\begin{proof}
In Lemma~\ref{lemma1}, $I$ is decomposed into four mutual-exclusive information capacity, $I_s$, $I_g$, $I_{sg}$, and $I_{t}$ in Eq.~\ref{infer5}. The detailed proof of Lemma~\ref{lemma1} can be found in Supplementary Material. 
\end{proof}


\subsection{Maximizing the Disentanglement among Spatial, Temporal and Graph Factors}
One of our goals is to maximize the disentanglement of spatial, temporal, and graph factors. So for example if a factor is merely related to spatial information, we do not want it to be explained by the spatial-graph joint factor $f_{sg}$. Analogously, if a factor is invariant to time, we do not want it to be explained by the time-variant factor $z_t$. However, this cannot be guaranteed by Equation \ref{infer5}, whose constraints can only enforce variable-level disentanglement within each type of factor instead of a maximized disentanglement across spatial, temporal, and graph factors. 

To address the above issue, we first re-interpret the constraints by information bottleneck theory \cite{burgess2018understanding}. The posterior distribution $q_{\phi}(z_{t}|S_t, G_t)$, $q_{\phi}(f_g|G_{1:T})$,$q_{\phi}(f_s|S_{1:T})$, and $q_{\phi}(f_{sg}|G_{1:T},S_{1:T})$ are interpreted as information bottleneck for the reconstruction task $E_{q_{\phi}(Z|G_{1:T},S_{1:T})}logp_{\theta}(S_{1:T}|z_{1:T},f_{s},f_{sg})$ and $E_{q_{\phi}(Z|G_{1:T},S_{1:T})}\log p_{\theta}(G_{1:T}|z_{1:T},f_{g},f_{sg})$. We propose that, by constraining the information flowing through each time-variant variable $z_{t}$ to be less than the information entropy of time-variant information $C_{t}$, namely $I_{t} \leq C_{t}$, $z_{t}$ will capture and only capture the time-variant information. We also propose that, by constraining the information flowing through the spatial-graph joint variable $f_{sg}$ to be less than the information entropy of the time-invariant correlated spatial-graph factor $C_{sg}$, namely $I_{sg} \leq C_{sg}$, $f_{sg}$ will only capture the time-invariant spatial-graph correlated factor. The new objective function is as follows:
\begin{align}
    \label{infer6}
    \nonumber
    \max_{\theta,\phi} \quad&\mathbb{E}_{S_{1:T}, G_{1:T}\sim \mathcal{D}}\mathbb{E}_{q_{\phi}(z_{1:T},F|S_{1:T}, G_{1:T})}\\\nonumber
    &\sum\nolimits_{t=1}^T[\log p_{\theta} (G_t|z_t,f_{g},f_{sg})+ \log p_{\theta}(S_t|z_t,f_{s},f_{sg})]  \\\nonumber
    \textrm{s.t.} &\quad \sum\nolimits_{t=1}^{T} D_{KL}( q_{\phi}(z_t|S_t,G_t)||p(z_t))<I_{t}\\\nonumber
    & \quad D_{KL}(q_{\phi}(f_{g}|G_{1:T})||p(f_{g}))<I_{g}\\\nonumber
    & \quad D_{KL}(q_{\phi}(f_{s}|S_{1:T})||p(f_{s}))<I_{s} \\\nonumber
    & \quad D_{KL}(q_{\phi}(f_{sg}|S_{1:T},G_{1:T})||p(f_{sg}))<I_{sg} \\
    & \quad I_{sg} \leq C_{sg}, \quad I_{t} \leq C_{t}.
\end{align}

This objective has the properties stated in Theorem~\ref{theorem1}.

\begin{theorem}
\label{theorem1}
The objective in Equation \ref{infer6} guarantees that $z_t$ captures and only captures the time-variant information while $f_{sg}$ captures and only captures the spatial-graph joint information.
\end{theorem}

\begin{proof}
The above theorem is proved based on the condition that (1) the sum of $I_s$, $I_g$, and $I_{sg}$ are large enough to contain the time-variant information, and $I_s$, $I_g$ are large enough to contain the time-invariant spatial-exclusive and graph-exclusive information, (2) $I_t \leq C_t$, and $I_{sg} \leq C_{sg}$. Due to the space limit,
the detailed proof is provided in Supplementary Material.
\end{proof}

\subsection{Spatiotemporal Graph Mutual Information Thresholding Algorithm}
Eq. \ref{infer6} is a challenging constrained nonconvex problem that also requires learning its hyperparameters of information bottleneck threshold $I_{sg}$ and $I_t$. This section proposes a novel algorithm along with its optimal condition analysis with respect to the information bottleneck threshold.

Given $I_s$ and $I_g$ are constants, the second and third constrain can be rewritten based on the Lagrangian algorithm under KKT condition~\cite{mangasarian1994nonlinear} as:
\begin{align}
   \label{eq:initial_objective2}
    \mathcal{R}_1 = \beta_2 &(D_{KL}(q_{\phi}(f_s|S_{1:T})||p(f_s)))
    \nonumber\\&+\beta_3 (D_{KL}(q_{\phi}(f_g|G_{1:T})||p(f_g)))
\end{align}
where the Lagrangian multipliers $\beta_2$ and $\beta_3$ are the regularization coefficients that control the capacity of the latent space information $f_s$ and $f_g$, respectively.

Next, since $I_t$ and $I_{sg}$ in the first constraint is a trainable parameter which ensures $I_{t} \leq C_{t}$ and $I_{sg} \leq C_{sg}$, it can be written as a Lagrangian under the KKT condition as
\begin{align}
   \label{eq:initial_objective1}
    &\mathcal{R}_2 = \beta_1 (\prod\nolimits_{t=1}^{T} D_{KL}(q_{\phi}(z_{t}|G_t,S_t)||p(z_t))-I_{t}) \\
    &\mathcal{R}_3 = \beta_4 ( D_{KL}(q_{\phi}(f_{sg}|S_{1:T},G_{1:T})||p(f_{sg}))-I_{sg}) \label{eq:initial_objective3}
\end{align}
Finally, we can optimize the overall objective as:
\begin{align}
   \small
   \label{eq:initial_objective4}
   \max_{\theta,\phi} \quad&\mathbb{E}_{S_{1:T}, G_{1:T}\sim \mathcal{D}}\mathbb{E}_{q_{\phi}(z_{1:T},F|S_{1:T}, G_{1:T})}\nonumber\\
   &\sum\nolimits_{t=1}^{T}[\log p_{\theta} (G_t|z_t,f_{g},f_{sg})+ \log p_{\theta}(S_t|z_t,f_{s_t},f_{sg})]\nonumber\\
   &-\mathcal{R}_1-\mathcal{R}_2-\mathcal{R}_3 \nonumber\\
   & \textrm{s.t.} \quad I_{sg} \leq C_{sg}, \quad I_{t} \leq C_{t}
\end{align}

It is very hard to directly optimize the above objective since $C_{t}$ and $C_{sg}$ are unknown. To circumvent the challenge, we propose a novel thresholding strategy consisting of two stages: the first stage is to optimize $I_{sg}$, the second stage is to optimize $I_{t}$, as detailed in Algorithm \ref{alg:A}. In short, we increase $I_{sg}$ by $\alpha$ in every $K$ until a stopping criteria is satisfied while keeping $I_{t}$ at a very low value (Lines 1-6 in Algorithm \ref{alg:A}). Then, we stop increasing $I_{sg}$ and increase $I_{t}$ by $\gamma$ every $J$ epoch until a stopping criterion is satisfied (Lines 8-14 in Algorithm \ref{alg:A}).

\begin{algorithm}[htb]
\caption{Information-iterative-thresholding algorithm}
\label{alg:A}
\begin{algorithmic}[1]
\REQUIRE The initialized parameter set $\mathcal{W}$; the initialized $I_{t}=\epsilon$ and $I_{sg}=\epsilon$ ($I_{t} \notin \mathcal{W}$ $I_{sg} \notin \mathcal{W}$ and $\epsilon$ is a very small number, e.g. $1\times 10^{-5}$); the increase step $\gamma$, $\alpha$ for optimizing $I_{t}$ and $I_{sg}$; the number of epochs $J, K$ of optimization for each updated $I_{t}$ and $I_{sg}$. 
\ENSURE The optimized parameter set $\mathcal{W}$.
\WHILE[\textit{stopping criterion for $I_{sg}$}]{$\mathcal{R}_3<0$} 
\STATE   $I_{sg}:=I_{sg}+\alpha$
\FOR[\textit{increase $I_{sg}$ every $K$ epoch}]{$epoch=1:K$} 
\STATE  Compute the gradient of $\mathcal{W}$ via backpropagation.
\STATE  Update $\mathcal{W}$ based on gradient with $I_{sg}$ and $I_{t}$ fixed.\\
\ENDFOR
\ENDWHILE
\WHILE[\textit{stopping criterion for $I_t$}]{$\mathcal{R}_2<0$} 
\STATE   $I_{t}:=I_{t}+\gamma$
\FOR[\textit{increase $I_{t}$ every $J$ epoch}]{$epoch=1:J$} 
\STATE  Compute the gradient of $\mathcal{W}$ via backpropagation.
\STATE  Update $\mathcal{W}$ based on gradient with $I_{t}$ and $I_{sg}$ fixed.\\
\ENDFOR
\ENDWHILE
\end{algorithmic}
\end{algorithm}
The proposed optimization strategy guarantees that $z_t$ captures and only captures the time-variant information while $f_{sg}$ captures and only captures the spatial-graph joint information based on the following theorem.

\begin{theorem}
The latent variable $z_t$ captures and only captures the time-variant information if $\mathcal{R}_2<0$ is satisfied. The latent variable $f_{sg}$ captures and only captures the time-invariant spatial-graph correlated information if $\mathcal{R}_3<0$ is satisfied.  
\label{theorem2}
\end{theorem}

\begin{proof}
Notably, at initial stage, $\mathcal{R}_3=0$ and  $\mathcal{R}_2=0$, we then gradually increase $I_t$ and $I_{sg}$, and at each step while well-trained,  $\prod\nolimits_{t=1}^{T} D_{KL}(q_{\phi}(z_{t}|G_t,S_t)||p(z_t))$ and $ D_{KL}(q_{\phi}(f_{sg}|G_{1:T},S_{1:T})||p(f_{sg}))$ will keep increasing to catch $I_t$ and $I_{sg}$.  When $\mathcal{R}_3<0$ and  $\mathcal{R}_2<0$, it indicates that the information that captured by $I_t$ and $I_{sg}$ do not increase anymore, namely $I_t=C_t$ and $I_{sg}=C_{sg}$. Thus, the whole optimization process can be stopped. During the whole process, the two constraint $I_{t} \leq C_{t}$ and $I_{sg} \leq C_{sg}$ are always satisfied, namely, $z_t$ always captures and only captures the time-variant information and $f_{sg}$ always captures and only captures the time-invariant spatial-graph correlated information. The detailed proof can be found in Supplementary Material. In practice, we set $\beta_1$, $\beta_2$, $\beta_3$, and $\beta_4$ as $1$ and the model is not sensitive to these parameters.
\end{proof}



Our model consists of four encoders which model $q_{\phi}(f_s|S_{1:T},G_{1:T})$, $q_{\phi}(f_g|S_{1:T},G_{1:T})$, $q_{\phi}(f_{sg}|S_{1:T},G_{1:T})$, and $q_{\phi}(z_t|S_{1:T},G_{1:T})$ respectively. There are also two decoders to model $p_{\theta} (G_t|z_t,f_{g},f_{sg})$ and $p_{\theta} (S_t|z_t,f_{g},f_{sg})$, respectively. Specifically, we utilize a typical graph convolution neural network to encode the graph factors and a typical convolution neural network for the spatial factors. For the spatial-graph correlated factors, we utilize a Spatial-Network Message Passing Neural Network (S-MPNN) \cite{guo2021spatial}, which considers both the spatial and graph information while passing messages. In terms of the temporal factors, we consider that could involve both spatial and graph variance, thus, we take another S-MPNN for the temporal factors. For decoders, we utilize a typical convolution neural network for the spatial factors, and a similar graph decoder proposed in NED-VAE \cite{guo2018deep} for the graph factors. Due to the space limits, the detailed description of the encoders and decoders, as well as the time complexity analysis are provided in the Supplementary Material.

\section{Experiments}
\label{sec:exp}

\small
\begin{table*}[htb]
     \centering
     \small
     \caption{The evaluation results for the generated spatiotemporal graphs for different datasets (\textit{KLD\_cls} refers to KLD of graph clustering coefficient. \textit{KLD\_ds} refers to KLD of graph density, \textit{KLD\_bet} refers to KLD of betweenness centrality, and \textit{KLD\_tcorr} refers to KLD of temporal correlation. (Best results are highlighted in bold, The KLD evaluation on the traffic dataset is not reported because the metrics are on graph topology, while the graph topology for the traffic dataset is unchanged.)}   
     \begin{adjustbox}{max width=\textwidth,height=0.6\columnwidth}
     \begin{tabular}{|c|c|rrrrrrr|r|}
         \hline
          Dataset&Method & Node &Spatial &Edge &KLD\_cls&KLD\_ds & KLD\_bet & KLD\_tcorr & AvgMI\\\hline
          \multirow{8}*{DWR Graph}&
          DSBM &N/A&N/A&54.95\%&0.90&1.10&0.63&0.73&N/A\\
          ~&GraphVAE &0.57&0.57&57.14\%&1.63&1.82&0.91&0.85&N/A\\
          ~&GraphRNN &N/A&N/A&55.24\%&1.97&2.50&1.00&1.35&N/A\\
          ~&beta-VAE&0.0012&0.0011&69.05\%&0.43&1.61&1.82&0.36&2.25\\
          ~&beta-TCVAE &0.0013&0.0012&69.04\%&0.47&1.37&1.56&0.08&2.33\\
          ~&NEND-IPVAE &0.016&0.0008&65.80\%&1.39&1.82&2.78&0.11&2.52\\
          ~&STGD-VAE &\textbf{0.0003}&\textbf{0.0001}&\textbf{69.99\%}&\textbf{0.14}&0.74&\textbf{0.40}&\textbf{0.03}&\textbf{2.03}\\
          ~&STGD-VAE-Dep &0.0191&0.0005&66.28\%&0.45&\textbf{0.55}&0.54&0.38&2.04\\\hline
          \multirow{8}*{DRG Graph}&
          DSBM &N/A&N/A&81.88\%&1.77&2.87&3.38&0.64&N/A\\
          ~&GraphVAE &0.56&0.74&85.75\%&4.46&2.65&1.60&3.08&N/A\\
          ~&GraphRNN &N/A&N/A&85.32\%&0.57&1.24&2.40&0.85&N/A\\
          ~&beta-VAE &0.0013&0.0017&91.75\%&0.34&1.24&1.47&2.15&2.29\\
          ~&beta-TCVAE &0.0018&0.0019&91.62\%&0.52&1.58&1.46&2.38&2.24\\
          ~&NED-IPVAE &0.0175&0.0018&89.84\%&0.37&1.05&1.72&0.23&2.42\\
          ~&STGD-VAE &\textbf{0.0004}&\textbf{0.0015}&\textbf{91.88\%}&\textbf{0.14}&0.72&0.28&\textbf{0.11}&\textbf{2.07}\\
          ~&STGD-VAE-Dep &0.0008&0.0017&91.28\%&\textbf{0.14}&\textbf{0.71}&\textbf{0.26}&1.67&2.08\\\hline
         \multirow{8}*{Protein} &
          DSBM &N/A&N/A&70.78\%&1.00&0.93&1.15&1.53&N/A\\
          ~&GraphVAE &N/A&553.82&62.54\%&1.26&1.44&1.48&1.90&N/A\\
          ~&GraphRNN &N/A&N/A&71.17\%&1.05&1.15&1.43&0.83&N/A\\
          ~&beta-VAE&N/A&52.74&85.58\%&0.16&\textbf{0.14}&0.46&0.61&1.04\\
          ~&beta-TCVAE &N/A&35.05&95.80\%&0.27&0.58&\textbf{0.34}&0.71&1.09\\
          ~&NED-IPVAE &N/A&36.12&92.48\%&1.08&0.79&0.44&2.64&1.15\\
          ~&STGD-VAE &N/A&28.77&\textbf{99.79\%}&0.33&0.21&0.53&\textbf{0.23}&\textbf{0.70}\\
          ~&STGD-VAE-Dep &N/A&\textbf{28.42}&96.79\%&\textbf{0.13}&0.54&1.55&0.24&0.76\\\hline
         \multirow{8}*{Traffic} &
          DSBM &N/A&N/A&N/A&N/A&N/A&N/A&N/A&N/A\\
          ~&GraphVAE &N/A&N/A&N/A&N/A&N/A&N/A&N/A&N/A\\
          ~&GraphRNN &N/A&N/A&N/A&N/A&N/A&N/A&N/A&N/A\\
          ~&beta-VAE &7.15&N/A&N/A&N/A&N/A&N/A&N/A&1.37\\
          ~&beta-TCVAE &8.50&N/A&N/A&N/A&N/A&N/A&N/A&1.18\\
          ~&NED-IPVAE &31.95&N/A&N/A&N/A&N/A&N/A&N/A&1.18\\
          ~&STGD-VAE &6.78&N/A&N/A&N/A&N/A&N/A&N/A&\textbf{0.69}\\
          ~&STGD-VAE-Dep &\textbf{5.13}&N/A&N/A&N/A&N/A&N/A&N/A&1.06\\\hline
     \end{tabular}
     \end{adjustbox}
     \label{tab:generation_results}
\end{table*}\normalsize

\begin{figure*}[htb]
\centering
\includegraphics[width=0.9\linewidth]{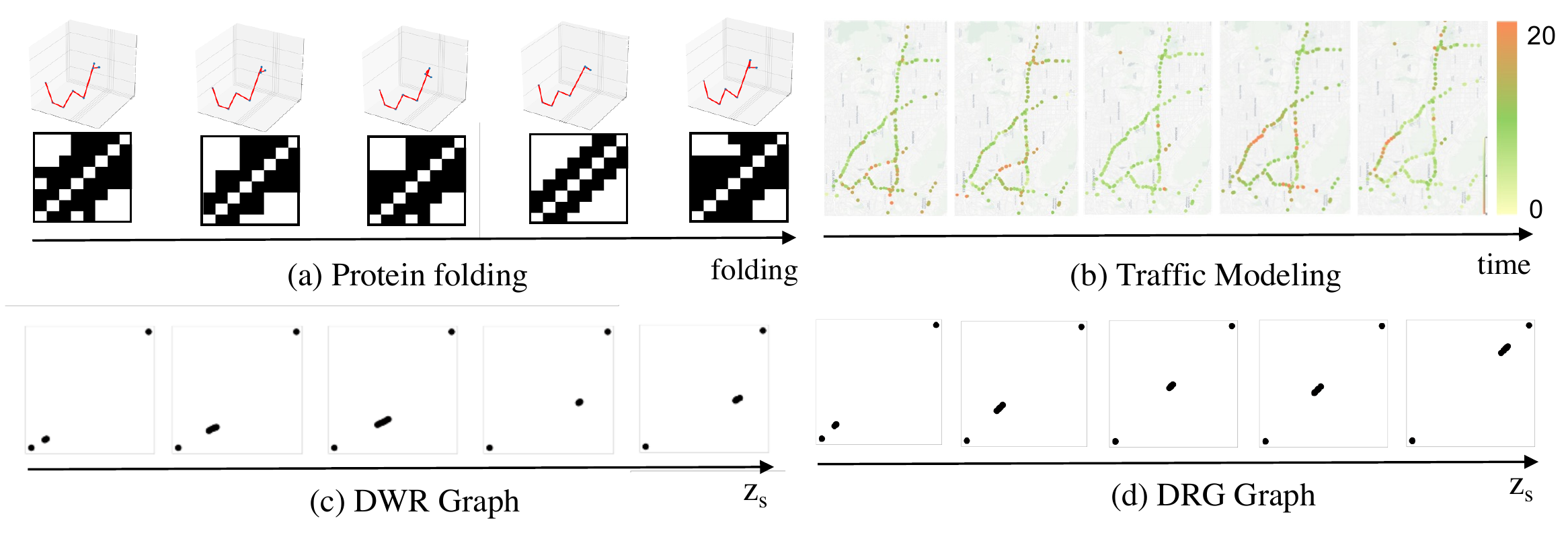}
\caption{Qualitative evaluation on two synthetic datasets and two real-world datasets.}
\label{fig:qual}
\end{figure*}

\subsection{Datasets}
\label{sec:dataset}
We validate the effectiveness of our proposed models on two synthetic datasets and two real-world datasets, (1) \textit{Dynamic Waxman Random Graphs}, (2) \textit{Dynamic Random Geometry Graphs}, (3) \textit{Protein Folding Dataset}, and (4) \textit{Traffic Dataset MERT-LA}~\cite{du2021graphgt}. The first two are well-known spatial network datasets \cite{bradonjic2007giant,waxman1988routing}, which randomly place nodes in a geometry and the edges are connected by predefined distance measures, with variances through the time dimension. The protein folding dataset consists of the folding steps of a protein with 8 amino acids \cite{guo2020interpretable}. The traffic dataset contains sequences of graphs which contains traffic speeds connected by 207 sensors \cite{jagadish2014big}. The details of generation process of the datasets can be found in Supplementary Material.

\subsection{Comparison Methods}

To validate the proposed models in spatiotemporal graph generation, despite that no previous deep models specially designed for the spatiotemporal graph generation task, we compare with some state-of-the-art graph generation model for generic graphs, including GraphRNN \cite{you2018graphrnn}, a graph generation method that utilizes recurrent neural network structure to model graphs as sequences of edges and nodes, GraphVAE \cite{kipf2016variational}, a variational auto-encoder based graph generative model focusing on handling small graphs, and a traditional algorithm DSBM \cite{xu2014dynamic} which utilizes stochastic block model for static graph generation and combines it with Markov chains to achieve temporal graph generation. For GraphRNN and GraphVAE, we train individual models for each time step respectively for generating temporal graph sequences.  

To validate the significance of our proposed disentanglement objective and optimization strategy for spatiotemporal graphs modeling, several SOTA disentanglement techniques are compared including beta-VAE \cite{higgins2016beta}, beta-TC-VAE \cite{chen2018isolating}, and NED-IPVAE \cite{guo2020interpretable}, where the architecture utilized is the same as our model, except with different disentanglement objectives. 

\subsection{Evaluation of the Performance in Spatiotemporal Graph Generation}
To validate the power of our proposed models to learn the representations of spatiotemporal graphs, we evaluate the reconstruction performance among different elements of a spatiotemporal graph, e.g. node, edge, and spatial locations by calculating the difference between the real ones and reconstructed ones. The mean square errors (MSE) is calculated for node attribute prediction and spatial prediction. The accuracy is used for evaluating edge connection prediction. To evaluate the generation performance, we select several common evaluation metrics to measure graph-structured data in multiple aspects, including graph statistics and temporal statistics. Specifically, we calculate a commonly used measurement for probability distributions, Kullback-Leibler Divergence (KLD), between the real training data distribution and the generated data distribution in terms of: (1) graph density, (2) average clustering coefficient, (3) betweenness centrality, and (4) temporal correlation in the spatiotemporal graphs.





The quantitative evaluation on the two synthetic datasets and two real-world datasets are reported in Table \ref{tab:generation_results}. First of all, for dynamic Waxman random graphs, STGD-VAE achieves the best overall evaluation which it not only reconstructs the node, spatial, and edge attributes very well, but also learns the underlying graph, temporal statistical distribution of the training set. To be specific, STGD-VAE outperforms the two disentangled VAE models, beta-VAE and beta-TCVAE by $1.3\%$ in reconstructing the graph edge connection. beta-VAE and beta-TCVAE fail to learn the exact underlying property statistical distribution of the graphs. We consider this as a benefit of the disentanglement of different types of features, which allows for clearer and better representations and results in a roughly $62.5\%$ improvement. 
For the dynamic random geometry graph dataset, the results are very similar to DWR Graph which mainly due to the similar way of generating the synthetic datasets, but with difference geometries. Obviously, our proposed STGD-VAE outperforms the best baseline model, beta-VAE, by at most $69.2\%$ in the reconstruction evaluation and roughly $58.8\%$ in learning the property distribution. 

In the two real-world datasets, the proposed STGD-VAE and STGD-VAE-Dep outperform other models in reconstructing the spatial locations of the nodes and predicting the contacts between residues by a large margin, which results in up to $19\%$ increase in predicting the spatial locations, and up to $4\%$ increase in contact prediction. However, in some of the graph statistics, our proposed models fail to beat all the baseline models because some of the graph statistics are not designed for the real-world protein structures. However, our proposed models still achieve comparable results in those evaluations. 
In the last traffic dataset, the graph topology and geometry stay the same, the only changing part is the node feature, which represent the traffic speeds. Our models achieve the best two in the speed prediction task and leave the third as highest as a $25.9\%$ gap.

\subsection{Evaluation of the Performance in Disentangled Representation learning}

\textbf{Quantitative Evaluation}. For quantitative evaluation, the typical \textbf{avgMI score}~\cite{locatello2019challenging} is utilized here to evaluate the power of the encoder to learn disentangled representations towards the predefined semantic factors by measuring the mutual distance between the latent variables and the semantic properties. Thus it is calculated based on the learn latent variables and the input graphs in the testing sets. As shown in Table~\ref{tab:generation_results}, for all the four datasets, beta-VAE, beta-TCVAE, and NED-IPVAE achieve similar results in terms of avgMI score. GraphRNN, GraphVAE, and DSBM are not capable of disentanglement learning, and thus, no score is reported on this evaluation metric. Between the two proposed models, STGD-VAE and STGD-VAE-Dep achieve comparable results in most of the cases, STGD-VAE is slightly better and have an up to $41.5\%$ improvement over the best baselines.

\textbf{Qualitative Evaluation}
As in the conventional qualitative evaluation in disentanglement representation learning \cite{chen2018isolating,higgins2016beta}, we change the value of one latent variable continuously while fixing the remaining variables to see the variation of the semantic factor it controls. In Fig. \ref{fig:qual}(a) and \ref{fig:qual}(b), we visualize the folding process of the protein structures and the traffic modeling process. We can observe that the residues on the right side are slightly folding up and moving towards left. For the traffic dataset, it is worth noting that the traffic speed is constantly changing in different time steps which reflects the real-time traffic situations. In Fig. \ref{fig:qual}(c) and \ref{fig:qual}(d), we also visualize the changes of the generated graphs when the latent factor $z_s$ of our STGD-VAE model change from $-5$ to $5$ in the dynamic Waxman random graph and the dynamic random geometry graph dataset, respectively. Clearly, the spatial location is changed accordingly, from the left-bottom corner to nearly the right-top corner, which shows that the latent variables learn and expose the semantic factors well.


\section{Conclusion}
\label{sec:conclusion}

In this paper, we introduce STGD-VAE and STGD-VAE-Dep, to the best of our knowledge, the first general deep generative model framework for spatiotemporal graphs. Specifically, we propose a new Bayesian model that factorizes spatiotemporal graphs into spatial, temporal, and graph factors as well as the factors that model the interactions among them. Moreover, a variational objective function and a new mutual information thresholding algorithm based on information bottleneck are proposed to maximize the disentanglement among the factors with theoretical guarantees. The comparison with six state-of-the-art deep generative models validates the superiority of our proposed models from multiple tasks, including, graph generation and disentangled representation learning. 


\section{Acknowledgement}
This work was supported by the National Science Foundation (NSF) Grant No. 1755850, No. 1841520, No. 2007716, No. 2007976, No. 1942594, No. 1907805, a Jeffress Memorial Trust Award, Amazon Research Award, NVIDIA GPU Grant, and NSF Computing Innovative Fellowship (Subaward No. 2021CIF-Emory-05).

\bibliography{STDVAE}



\end{document}